\documentclass[letterpaper, 10 pt, conference]{ieeeconf}  

\IEEEoverridecommandlockouts                              

\usepackage{graphics} 
\usepackage{epsfig} 
\usepackage{mathptmx} 
\usepackage{times} 
\usepackage{amsmath} 
\usepackage{amssymb}  

\usepackage{gensymb}
\usepackage{cite}
\usepackage{algorithmic}
\usepackage{textcomp}
\usepackage{scrextend}
\usepackage{changepage}
\usepackage{multirow}
\usepackage{amsfonts}
\usepackage{booktabs}
\usepackage{mathtools}
\usepackage{xcolor}
\usepackage{float}
\usepackage{makecell}

\usepackage[font={footnotesize}]{caption}

\usepackage{hyperref}

\usepackage[linesnumbered, ruled]{algorithm2e}

\DeclarePairedDelimiter\norm{\lVert}{\rVert}

\title{\LARGE \bf
Put the Bear on the Chair! Intelligent Robot Interaction \\ with Previously Unseen Chairs via Robot Imagination
}

\author{Hongtao Wu, Xin Meng, Sipu Ruan, Gregory S. Chirikjian
\thanks{This work was supported by NUS Startup grants R-265-000-665-133, R-265-000-665-731, Faculty Board account C-265-000-071-001, and the National Research Foundation, Singapore under its Medium Sized Center for Advanced Robotics Technology Innovation (CARTIN) R-261-521-002-592.}
\thanks{H. Wu, X. Meng, S. Ruan, and G. S. Chirikjian are with the Department of Mechanical Engineering,
        National University of Singapore, Singapore.
        H. Wu is also with the Laboratory for Computational Sensing and Robotics (LCSR), Johns Hopkins University, Baltimore, MD 21218, USA.
        Address all correspondence to G. S. Chirikjian: {\tt\small mpegre@nus.edu.sg}
    }
}

\begin{document}

\maketitle
\thispagestyle{empty}
\pagestyle{empty}

\begin{abstract}

In this paper, we study the problem of autonomously seating a teddy bear on a previously unseen chair.
To achieve this goal, we present a novel method for robots to imagine the sitting pose of the bear by physically simulating a virtual humanoid agent sitting on the chair.
We also develop a robotic system which leverages motion planning to plan SE(2) motions for a humanoid robot to walk to the chair and whole-body motions to put the bear on it.
Furthermore, to cope with cases where the chair is not in an accessible pose for placing the bear, a human assistance module is introduced for a human to follow language instructions given by the robot to rotate the chair and help make the chair accessible.
We implement our method with a robot arm and a humanoid robot.
We calibrate the proposed system with 3 chairs and test on 12 previously unseen chairs in both accessible and inaccessible poses extensively.
Results show that our method enables the robot to autonomously seat the teddy bear on the 12 previously unseen chairs with a very high success rate.
The human assistance module is also shown to be very effective in changing the accessibility of the chair.
Video demos and more details are available at \url{https://chirikjianlab.github.io/putbearonchair/}.

\end{abstract}

\section{INTRODUCTION}
\label{sec:introduction}
Object affordances describe potential interactions with an object \cite{gibson1979ecological}. 
They are related to the object functionality.
For example, a chair is able to afford the functionality of sitting.
Thus, it possesses the sitting affordance.
For robots operating in unstructured environments and interacting with previously unseen objects, \textit{e.g.}, personal robots, the understanding of object affordances is very desirable \cite{bogoni1995interactive}.
It allows robots to understand how to interact with them efficiently and intelligently.
Reasoning about object affordances can also help robots discover the potential of an object to afford a functionality, despite it not being a typical instance of the object class to which the functionality is related (\textit{e.g.}, the \textit{improvised} chair assembled with books and boxes in Fig. \ref{fig:4}(d)).
Independently identifying affordances represents a higher level of robot intelligence.

\begin{figure}[t]
    \centering
    \includegraphics[width=0.9\columnwidth]{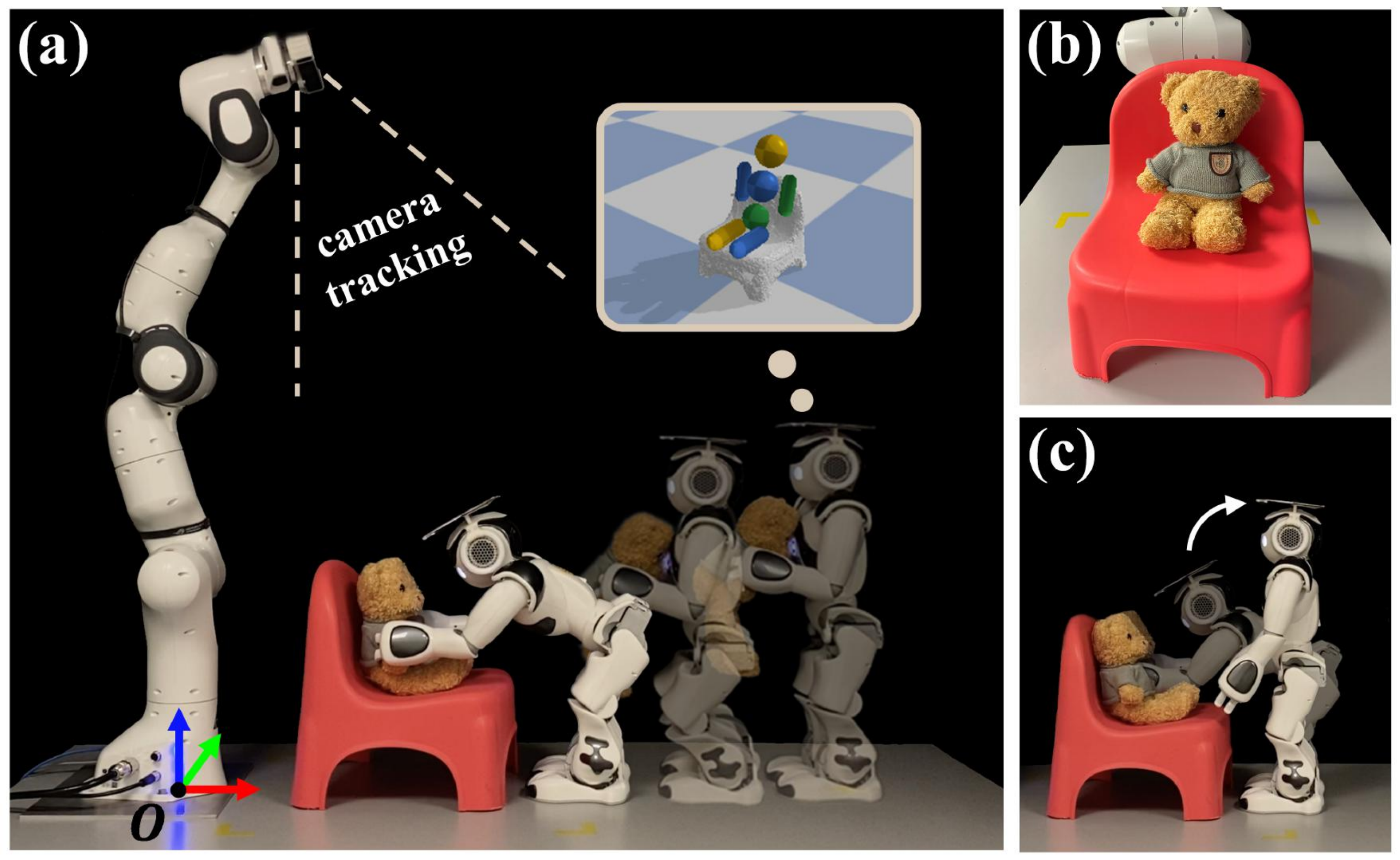}
    \caption{\textbf{Overview.} 
    (a) The robot imagines the sitting pose for a previously unseen chair and seats the bear on the chair. 
    A robot arm with a mounted depth camera scans the chair and tracks the chair and the humanoid robot. 
    The coordinate frame at the lower left corner indicates the base frame of the robot arm which is used as the world frame throughout the paper.
    (b) Result of seating the bear on the chair. 
    (c) The humanoid robot retrieves its hands and uprights its body after seating the bear on the chair.}
    \label{fig:1}
    \vspace{-0.6cm}
\end{figure}

To address the problem of affordance reasoning, we introduced the \textit{interaction-based definition (IBD)} which defines objects from the perspective of interactions instead of appearances in our previous work \cite{wu2020chair}.
In particular, IBD defines a chair as ``an object which can be stably placed on a flat horizontal surface in such a way that a typical human is able to sit (to adopt or rest in a posture in which the body is supported on the buttocks  and  thighs  and  the  torso  is  more  or less upright) on it stably above the ground".

\begin{figure*}
    \centering
    \includegraphics[width=1.5\columnwidth]{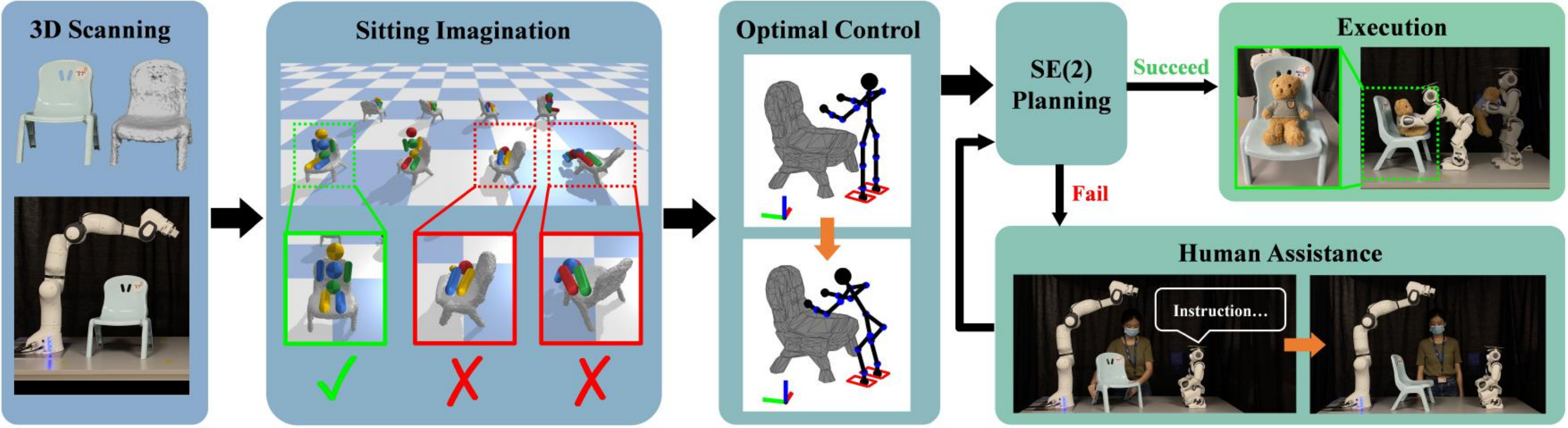}
    \caption{\textbf{Pipeline}. The chair is randomly placed in front of the robot arm for 3D scanning which reconstructs the 3D model of the chair. 
    The three snapshots in the sitting imagination show the results of three sitting trials. 
    Only the left one with a check is classified as a \textit{correct} sitting. 
    The instruction given by the robot in the human assistance module is "Please rotate the chair about the vertical axis counterclockwise for 90 degrees!"}
    \label{fig:2}
    \vspace{-0.7cm}
\end{figure*}

In this paper, we go beyond our previous work.
We propose a novel method for robots to imagine how to sit on a previously unseen chair, \textit{i.e.}, the sitting pose for the chair.
We develop a robotic system, with a Franka Emika Panda robot arm and a NAO humanoid robot, to actualize the understanding of the sitting affordance by autonomously seating a teddy bear on the chair (Fig. \ref{fig:1}).
To ``put the bear on the chair", the robot needs to carry the bear to the chair and seat the bear on it.
To accomplish this task, the robot first needs to understand the sitting affordance of the chair, \textit{i.e.}, how the chair can be sat.
Also, the robot needs to plan a whole-body motion to seat the bear on the chair.
Moreover, if the chair is in an inaccessible pose (\textit{e.g.}, when the chair is facing a wall), the robot should be able to reason the accessibility and understand how to make it accessible.
In this paper, we introduce a simple but efficient human assistance module with closed-loop visual feedback to enable the robot to instruct a human to rotate the chair and make it accessible for seating the bear.
Fig. \ref{fig:2} shows the pipeline of our method.
Results show that our method enables the robot to autonomously seat the bear on 12 previously unseen chairs with diverse shapes and appearances in 72 trials, including both accessible and inaccessible chair poses, with a success rate of $94.4\%$.
Also, our human assistance module is shown to be very effective. 
It successfully changes the accessibility of the chair for seating the bear with a $100\%$ success rate in 36 trials with inaccessible poses.
To our knowledge, this work is the first to physically seat an agent on a previously unseen chair in the real world.
The aim of this work is to examine how well the robots are able to accomplish this challenging task by leveraging different components of robotics.
We envision promising future applications of our method for robots operating in household environments and interacting with humans.

This work differs from our previous work \cite{wu2020chair} in various ways.
Rather than classifying whether a previously unseen object is a chair or not, in this work, we assume the given object is known as a chair \textit{a priori}, as such can be determined from \cite{wu2020chair}.
Also, unlike \cite{wu2020chair} which focuses on perceiving the sitting affordance without real-robot experiments, this work goes beyond by actualizing the understanding of affordances with physical experiments.
The main contributions of the paper are as follows:
\begin{itemize}
    \item A method for robots to imagine the sitting pose for a previously unseen chair.
    \item A human assistance module to enable the robot to reason about the accessibility of a chair and interact with a human to change the accessibility if necessary. 
    \item A robotic system which is able to autonomously seat a teddy bear on a previously unseen chair for sitting.
\end{itemize}

\section{RELATED WORK}
\label{sec:related work}
\textbf{Object Affordance of Chairs.}
The class of chairs is an important object class in human life.
The exploration of sitting affordance has become popular in recent decades \cite{hinkle2013predicting, grabner2011makes, seib2016detecting, wu2020chair, bar2006functional}.
Hinkle and Olson \cite{hinkle2013predicting} simulate dropping spheres onto objects and classify objects into containers, chairs, and tables based on the final configuration of spheres.
Grabner \textit{et al.} \cite{grabner2011makes} fit a humanoid mesh onto an object mesh and exploit the vertex distance and the triangle intersection to evaluate the object affordance as a chair for chair classification.
Instead of object classification, we use object affordances to understand the interaction with a chair.

\textbf{Affordance Reasoning.}
There is a growing interest in reasoning object affordances in the field of computer vision \cite{myers2015affordance, sawatzky2017weakly, roy2016multi, ruiz2020geometric, zhu2014reasoning, ho1987representing, desai2013predicting, bogoni1995interactive, bansal2020detecting} and robotics \cite{stoytchev2005behavior, do2018affordancenet, chu2019learning, wu2020can, abelha2017learning, piyathilaka2015affordance, bajcsy1988active, bogoni1995interactive, kjellstrom2011visual}.
Stoytchev \cite{stoytchev2005behavior} introduces an approach to ground tool affordances via dynamically applying different behaviors from a behavioral repertoire.
\cite{do2018affordancenet, chu2019learning, sawatzky2017weakly, roy2016multi, desai2013predicting} use convolutional neural networks (CNNs) to detect regions of affordance in an image.
Ruiz and Mayol-Cuevas \cite{ruiz2020geometric} predict affordance candidate locations in environments via the interaction tensor.
In contrast to detecting different types of affordances, we focus on understanding the sitting affordance and use it for real robot experiments.

\textbf{Physics Reasoning.}
Reasoning of physics has also been introduced to the study of object affordances \cite{wu2020chair, wu2020can, battaglia2013simulation, zhu2015understanding, zhu2016inferring, kunze2017envisioning}.
Battaglia \textit{et al.} \cite{battaglia2013simulation} introduce the concept of an intuitive physics engine which simulates physics to explain human physical perception of the world.
Zhu \textit{et al.} \cite{zhu2016inferring} employ physical simulations to infer forces exerted on humans and learn human utilities from videos.
The digital twin \cite{boschert2016digital} makes use of physical simulations to optimize operations and predict failures in manufacturing.
We do not infer physics or use it to predict outcomes, but exploit physics to imagine interactions with objects to perceive object affordances.

\section{METHODS}
\label{sec:methods}
\subsection{Problem Formulation}
\begin{figure}[t]
    \centering
    \includegraphics[width=0.87\columnwidth]{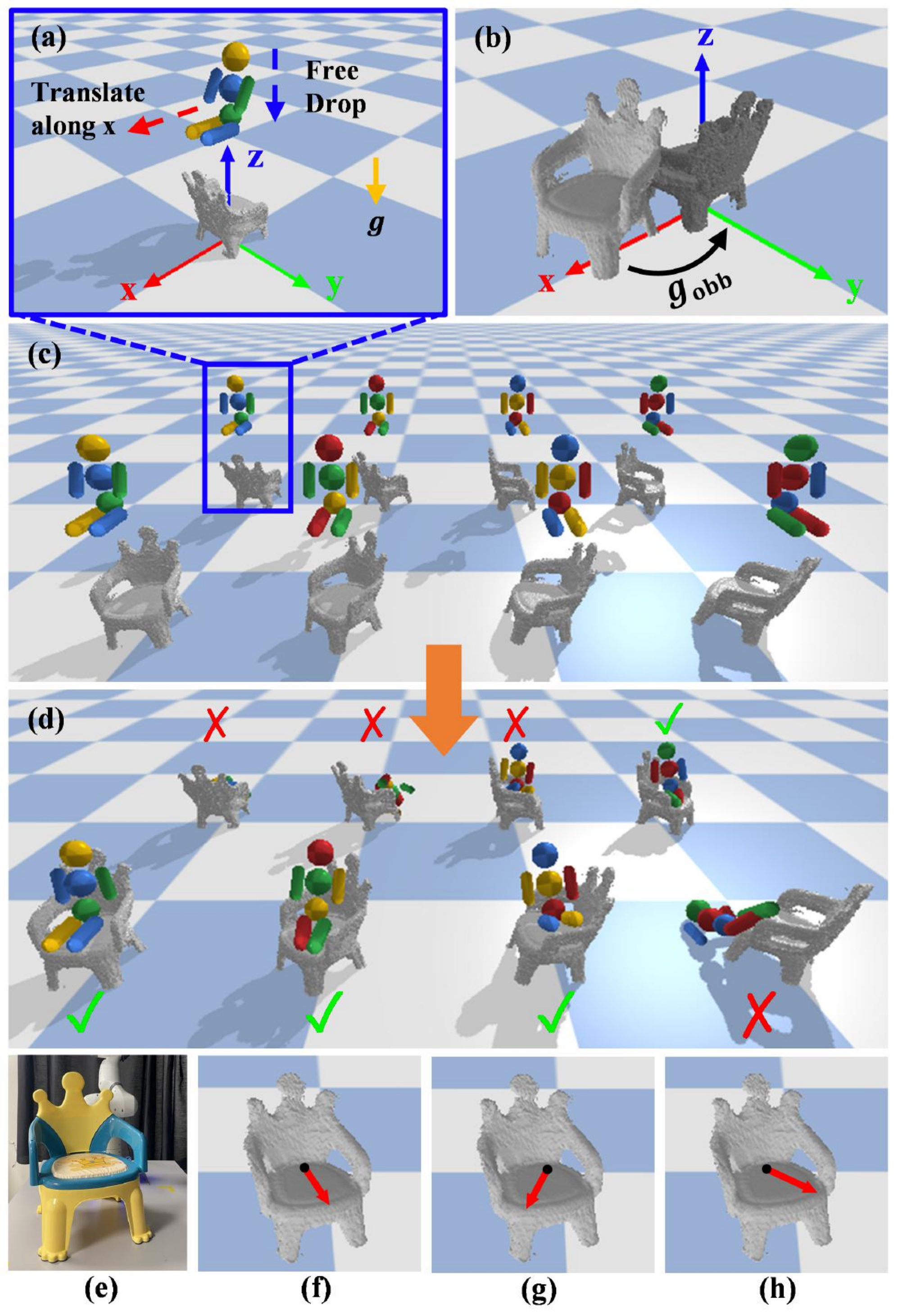}
    \caption{
    \textbf{Sitting imagination.}
    (a) Imagination setting.
    (b) OBB transformation $g_{obb}$. 
    (c) and (d) show the snapshots at the beginning and the end of the sitting imagination, respectively. 
    The checks and crosses in (d) indicate whether the sitting is \textit{correct} or not. 
    (e) shows the chair used for imagination in (c) and (d). 
    (f), (g), and (h) show the sitting poses obtained from three rotations $\alpha_{i}$.
    $N_{correct}=3$ for all of them.
    The black dot and the red arrow indicate the sitting position $\mathbf{p}$ and the yaw angle $\gamma$, respectively.
    (f) has the smallest value of $JL$ and thus is selected as the imagined sitting pose for the robot to seat the bear on the chair.}
    \label{fig:3}
    \vspace{-0.6cm}
\end{figure}

Throughout the paper, we assume the given unseen chair is upright and the back of the agent is supported while sitting.
The virtual agent in the imagination is an articulated humanoid body (Fig. \ref{fig:3}(a)).
To seat the agent on the chair, we need to know 1) the sitting pose $g = (R, \mathbf{p}) \in SE(3)$ where $R \in SO(3)$ and $\mathbf{p} \in \mathbb{R}^{3}$ specify the rotation and position of the agent's base link in the world frame and 2) the joint angles of the agent.
However, the real agent is a teddy bear of which the joints have large damping coefficients.
The joints can be considered fixed and the joint angles are already close to those of a sitting configuration (Fig. \ref{fig:1}(b)).
Thus, in this paper, we simplify the problem to just finding $g$.

According to the interaction-based definition (IBD) of chairs (Sec. \ref{sec:introduction}), the torso is more or less upright when sitting.
Thus, we further simplify the problem by restricting $R = R_{z}(\gamma)R_{0}$.
$R_{z}(\gamma)$ denotes the rotation about the z-axis of the world frame and $\gamma$ is the yaw angle of the base link.
$R_{0}$ denotes the initial rotation which sets the agent to an upright sitting configuration with its face facing towards the x-axis of the world frame.
That is, given an unseen chair, the problem becomes finding the position $\mathbf{p}$ and the yaw angle $\gamma$ of the base link in the world frame for sitting.
We denote the direction indicated by $\gamma$ in the xy-plane as the \textit{sitting direction}.
The base link of the agent is the pelvis link.

\subsection{Revisit of Sitting Affordance Model (SAM)}
\label{subsec:SAM}
In sitting imagination (Sec. \ref{subsec:sitting imagination}), we simulate a virtual humanoid agent sitting on the chair in physical simulations.
The resultant configuration of the agent at the end of the simulation is denoted as $C_{{res}}$.
The sitting affordance model (SAM) \cite{wu2020chair} evaluates whether $C_{{res}}$ is a \textit{correct} sitting by comparing it with a key configuration $C_{{key}}$.
We briefly review SAM here and further details can be found in \cite{wu2020chair}.

The evaluation is based on four criteria.
\textbf{1) Joint Angle Score.}
The joint angles of a configuration $C$ can be described with a vector $\boldsymbol{\theta} \in \mathbb{R}^{n}$ where $n$ denotes the total number of the agent's joints.
The joint angle score is defined as $J = \sum_{i} w_{J}^{i}|\theta_{{res}}^{i} - \theta_{{key}}^{i}|$.
$w_{J}^{i}$ denotes the weight of the $i$-th joints.
$\theta_{{res}}^{i}$ ($\theta_{{key}}^{i}$) is the $i$-th element of $\boldsymbol{\theta}_{{res}}$ ($\boldsymbol{\theta}_{{key}}$).
Lower is better.
\textbf{2) Link Rotation Score.}
According to IBD, the torso is more or less upright. 
Thus, SAM considers the link rotation in the world frame when evaluating $C_{{res}}$. 
The link rotation score is defined as $L =\sum_{i} w_{L}^{i}( 1 - \mathbf{z}_{{res}}^{i} \cdot \mathbf{z}_{{key}}^{i})$.
$w_{L}^{i}$ is the weight for the $i$-th link.
$\mathbf{z}_{{res}}^{i}$ ($\mathbf{z}_{{key}}^{i}$) is the z-axis unit vector of the $i$-th link frame in $C_{{res}}$ ($C_{{key}}$).
Lower is better.
\textbf{3) Sitting Height.}
Sitting height $H$ is also an important factor in sitting. SAM takes into account $H$ in the evaluation of $C_{{res}}$.
\textbf{4) Contact.}
SAM counts the number of contact points of the agent's links which can be described with a vector $\mathbf{T} \in \mathbb{R}^{m}$ where $m$ denotes the total number of links.

$C_{{res}}$ is classified as a correct sitting if the following are all satisfied: $J < J_{\textrm{max}}$, $L < L_{\textrm{max}}$, $H \in (H_{\textrm{min}}, H_{\textrm{max}})$, $\phi(\textbf{T}) > 0$.
$J_{\textrm{max}}$, $L_{\textrm{max}}$, $H_{\textrm{min}}$, and $H_{\textrm{max}}$ are four thresholds.
$\phi(\cdot): \mathbb{R}^{m} \rightarrow \{0, 1\}$ is a binary function which outputs 1 if 1) the total number of contact points is larger than some thresholds and 2) the number of contact points for lower and upper body links are both larger than zeros, and 0 if otherwise.

\subsection{Sitting Imagination}
\label{subsec:sitting imagination}
We physically simulate the agent sitting on the chair in different rotations $R_{{chr}} \in SO(3)$ to find $\mathbf{p}$ and $\gamma$ for sitting (Fig. \ref{fig:3}).
Given the 3D model of the chair, we first compute the minimum volume oriented bounding box (OBB) \cite{trimesh}.
As in \cite{wu2020chair}, we then apply a rigid body transformation $g_{{obb}} = (R_{{obb}}, \mathbf{p}_{{obb}}) \in SE(3)$ to the model (Fig. \ref{fig:3}(b)).
$\mathbf{p}_{{obb}}$ horizontally translates the chair to align the OBB center with the origin of the world frame in the xy-plane.
$R_{{obb}} = R_{z}(\gamma_{{obb}})$ rotates the chair about the z-axis to align the OBB with the coordinate system of the world frame.
We apply $g_{{obb}}$ because we notice that the back of the chair is heuristically coincident with one of the OBB faces which benefits the finding of correct sittings in the imagination.
After applying $g_{obb}$, we attach the world frame as the body frame of the chair.

The rotation of the chair $R_{chr}$ in the simulation is enumerated by setting $R_{{chr}} = R_{z}(\alpha_{i})R_{{obb}}$  and $\alpha_{i} = i\pi / 4, i = 0, 1, 2, ..., 7$ (Fig. \ref{fig:3}(c)).
Note that $R_{{chr}} = R_{z}(\alpha_{i})R_{{obb}} = R_{z}(\alpha_{i} + \gamma_{{obb}})$ is a rotation about the z-axis in the world frame.
We drop the agent from above the chair to simulate sitting (Fig. \ref{fig:3}(d)).
Before the drop, the agent is set to a pre-sitting configuration facing the x-axis (Fig. \ref{fig:3}(a)).
The base link of the agent is placed on a plane 15cm above the chair OBB.
For each rotation, we first sample three positions on the plane to drop: the origin and two positions with a translation of $\pm L_{sit}$ along the x-axis from the origin, respectively.
$L_{sit}$ is scaled linearly with respect to the size of the OBB.
If no more than one correct sitting is found for all rotations, four extra positions are sampled to drop: positions with $\pm 2L_{sit}$ and $\pm 3L_{sit}$ translations along the x-axis from the origin, respectively.
The reason we start sampling drops close to the origin of the plane, which is aligned with the center of the OBB horizontally, is that most chairs have their seats positioned close to the OBB center.
However, the seats for some chairs are closed to the OBB peripheral.
Thus, if not enough correct sittings can be found, our search expands towards the peripheral.
In total, we simulate 24 drops if no extra drops are needed and 56 drops if otherwise.

The rotation with the largest number of correct sittings $N$ is selected as the best rotation $\alpha^{*}$ for sitting.
If more than one rotation has the largest $N$ (Fig. \ref{fig:3}(f)(g)(h)), we select the one with the smallest averaged value of $JL$ (lower is better).
The sitting pose $g=(R, \mathbf{p})$ in the world frame is:
\begin{gather}
    \vspace{-0.5cm}
    \mathbf{p} = g_{obb}^{-1} \circ \overline{\mathbf{p}'} \\
    R = R_{z}(\gamma) R_{0}, \quad \gamma = \overline{\gamma'} - \gamma_{obb}
    \vspace{-0.5cm}
\end{gather}
where $g \circ \mathbf{p}' = R\mathbf{p}' + \mathbf{p}$ and the inverse $g^{-1} \circ \mathbf{p}' = R^{-1}\mathbf{p}'-R^{-1}\mathbf{p}$; $\overline{\mathbf{p}'}$ and $\overline{\gamma'}$ are the weighted average of the agent's position and yaw angle relative to the chair frame of all the correct sittings with the chair rotated by $\alpha^{*}$.
The weight is the reciprocal value of $JL$ of the sitting.

\begin{figure}[t]
    \centering
    \includegraphics[width=1\columnwidth]{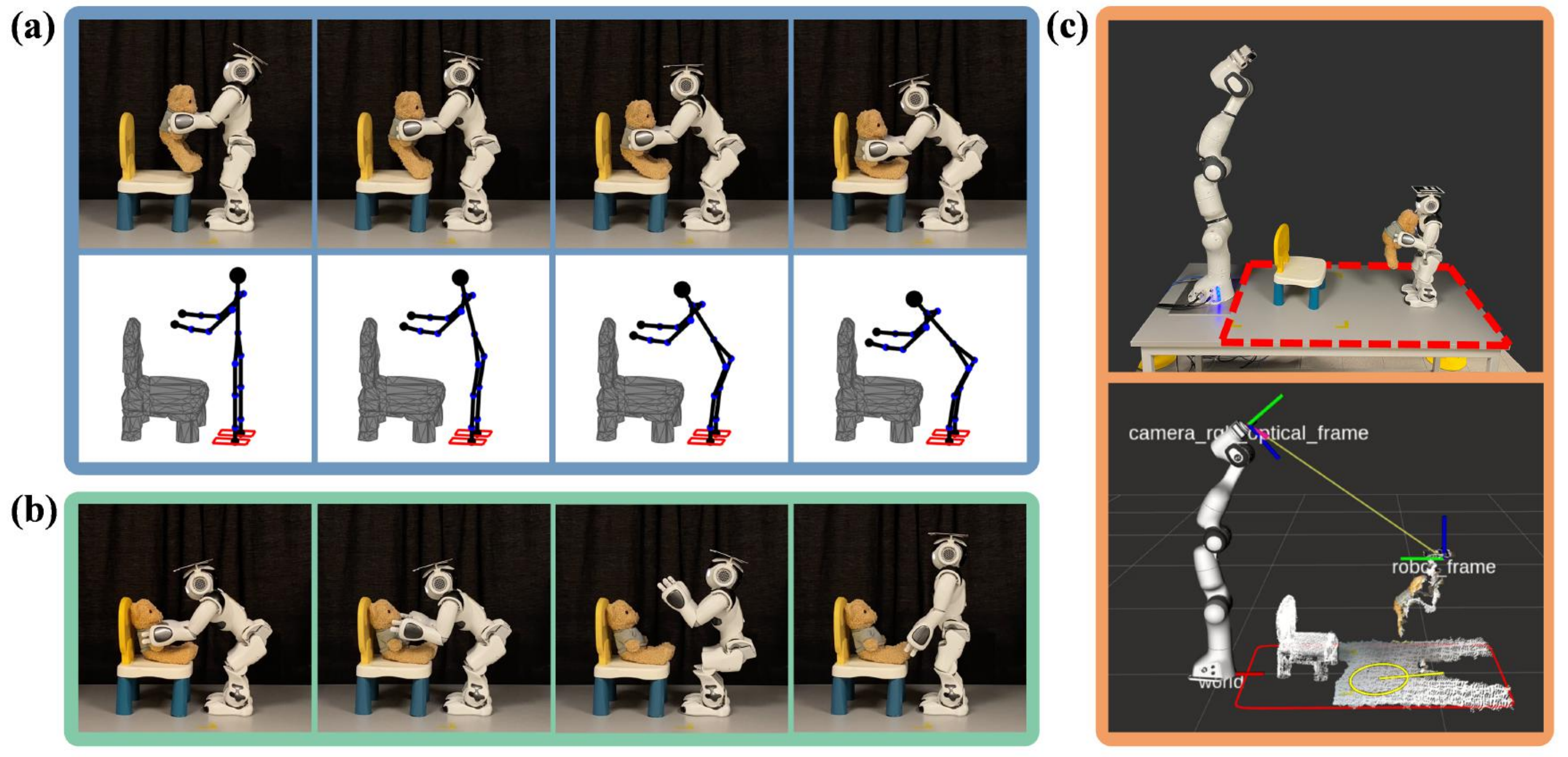}
    \caption{
    \textbf{Motion Planning.}
    (a) The motion of seating the bear on the chair.
    (b) After seating the bear, the robot releases the bear by opening its hands, retrieves the hands, and uprights the body.
    (c) The setting of the $SE(2)$ planning in real and rviz environment. 
    The red dash lines in the top figure indicate the planning arena in the real world.
    The yellow ellipsoid in the bottom figure shows the $SE(2)$ goal.
    }
    \label{fig:5}
    \vspace{-0.7cm}
\end{figure}

\subsection{Motion Planning}
\label{sec:motion planning}

We assume the motion of seating the bear is quasi-static.
Due to the complexity of the motion constraints, we formulate the planning of this motion as a trajectory optimization problem \cite{han2020can}:
\begin{align}
    & \underset{\mathrm{\mathbf{q}}, \dot{\mathrm{\mathbf{q}}}}{\text{minimize}}
    & & J = \sum_{k=1}^{N}L(\mathrm{\mathbf{q}}(k)) \\
    & \text{subject to}
    && \text{Kinematics constraints} \label{eqn:9} \\
    &&& \text{Stability constraints} \label{eqn:10} \\
    &&& \text{Collision constraints} \label{eqn:11}
\end{align}
where $L(\mathrm{\mathbf{q}}(k)) = 1/2\norm{\mathrm{\mathbf{q}}(k) - \mathrm{\mathbf{q}}_{{goal}}}_{Q}^{2}$.
Kinematics constrains describe the state transition and limit the magnitude of joints and control inputs.
Stability constrains confine the horizontal projection of the robot center of mass (COM) to the supporting polygon $\mathbf{S}$ formed by the feet.
Collision constraints ensure that the trajectory is collision-free.

The initial configuration of the robot is a pre-defined standing posture (Fig. \ref{fig:5}(a)).
The goal configuration $\mathrm{\mathbf{q}}_{{goal}}$ is generated via a constrained optimization \cite{han2020can}:
\begin{align}
    & \underset{\mathrm{\mathbf{q}}}{\text{minimize}}
    & & w_{1}d_{\mathrm{COM}}(\mathrm{\mathbf{q}}) + w_{2}\mathrm{Torso}(\mathrm{\mathbf{q}}) \label{eqn:12}\\ 
    & \text{subject to}
    & & \text{Kinematics constraints} \label{eqn:14} \\
    &&& \text{Stability constraints} \label{eqn:15} \\
    &&& \text{Collision constraints} \label{eqn:16}
\end{align}
The cost (Eq. \eqref{eqn:12}) aims to minimize 1) $d_{\mathrm{COM}}(\mathrm{\mathbf{q}})$ which is the distance between the COM horizontal projection and the center of the supporting polygon $\mathbf{S}$ and 2) $\mathrm{Torso}(\mathrm{\mathbf{q}})$ which is the bending angle of the torso.
A less bending torso exerts less torque on the motors, making uprighting the body easier after seating the bear.
Kinematics constraints ensure that the bear sits in the imagined sitting pose $g$ when the robot is in the goal configuration $\mathbf{q}_{goal}$.

We use RRTConnect \cite{kuffner2000rrt} in the OMPL library \cite{sucan2012the-open-motion-planning-library} to plan an $SE(2)$ trajectory for the robot to carry the bear to the chair.
The robot is encapsulated with an ellipsoid for collision checking with the FCL library \cite{pan2012fcl}.
The setting is shown in Fig. \ref{fig:5}(c).

\section{SYSTEM PIPELINE}
\label{sec:pipeline}
Fig. \ref{fig:2} shows the pipeline of our method.
The robot arm first scans and reconstructs the 3D model of the chair (Sec. \ref{sec:scanning}).
The sitting imagination is then conducted to find the imagined sitting pose $g$ (Sec. \ref{subsec:sitting imagination}).
With $g$, we first determine the $SE(2)$ goal pose $s_{goal}$ for the NAO to walk to and place the bear.
The rotation of $s_{goal}$ is set such that the NAO faces the opposite of the sitting direction indicated by $\gamma$.
The position of $s_{goal}$ is on a horizontal ray which originates from the projection of $\mathbf{p}$ on the xy-plane and points towards the sitting direction.
It is initially set such that the NAO is $l_{init}$ away from $\mathbf{p}$.
If the NAO is in collision with the chair, we move it along the ray until it is collision-free.
After that, if the distance between the NAO and $\mathbf{p}$ is too large, we move $\mathbf{p}$ horizontally along the ray and make it closer to the robot due to the robot workspace constraint.
We then pre-plan the motion to seat the bear and check the validity of $s_{goal}$.
To reduce the planning time, the motion is simplified as a bilaterally symmetric motion -- the motion of the left-half body is symmetric to that of the right-half.
The bear is held in hands facing the sitting direction at the beginning of the motion (Fig. \ref{fig:5}(a)).
We restrict the motion to be an $SE(2)$ motion, in which only the pitch joints are activated, to maintain the facing of the bear throughout the motion.

The $SE(2)$ trajectory to walk to $s_{goal}$ is then planned.
If $s_{goal}$ is out of the planning arena or blocked by obstacles, no plans will be made.
In this case, the NAO gives a language instruction to instruct a human to assist with rotating the chair about the vertical axis such that the sitting direction points towards the NAO where there are no obstacles in between (Fig. \ref{fig:2}).
The instruction is generated from a template: ``Please rotate the chair about the vertical axis $\tt{direction}$ for $\tt{rotation\_angle}$ degrees".
$\tt{direction} \in \{\tt{clockwise}, \tt{counterclockwise}\}$.
$\tt{rotation\_angle} \in \{\beta | \beta=i * 30, i=0, 1, 2, ..., 6\}$ is the multiple of 30 degrees closest to the precise rotation angle such that the sitting direction is pointing towards the NAO.
We use multiples of 30 degrees instead of the precise angle because it is easier for humans to understand.
The pose of the chair is tracked by iterative closest point (ICP) \cite{besl1992ICP}.
The transformation of the chair in the human rotation is denoted as $g_{rot}$.
The imagined sitting pose $g$ and the $SE(2)$ goal $s_{goal}$ are updated and transformed by $g_{rot}$ after the rotation.
The robot then tries to plan a trajectory to the updated $s_{goal}$.
This process is repeated until a valid $SE(2)$ trajectory is found.
In practice, we regard the trial as a failure if no $SE(2)$ trajectories can be found after three rotations.
When a valid $SE(2)$ trajectory is found, the NAO is controlled to walk and follow the waypoints along the trajectory via a PID controller.
The walking motion is controlled by the NAOqi SDK\footnote{https://developer.softbankrobotics.com/nao6/naoqi-developer-guide/naoqi-apis}.
The bear is passed to the NAO manually before it starts walking.
When the NAO arrives at $s_{goal}$, a whole-body motion is planned to place the bear based on the robot current pose.
Finally, the robot executes the motion, releases the bear, and uprights its body (Fig. \ref{fig:5}(b)).

\section{EXPERIMENTS}
\label{sec:experiments}
Fig. \ref{fig:1}(a) shows the experiment setup.
A PrimeSense Carmine 1.09 RGB-D camera is mounted on the robot arm.
Besides scanning the chair, the camera is also used for tracking the chair and the NAO.
The NAO is tracked with an ArUco tag attached on top of its head.

\begin{figure}[t]
    \centering
    \includegraphics[width=0.95\columnwidth]{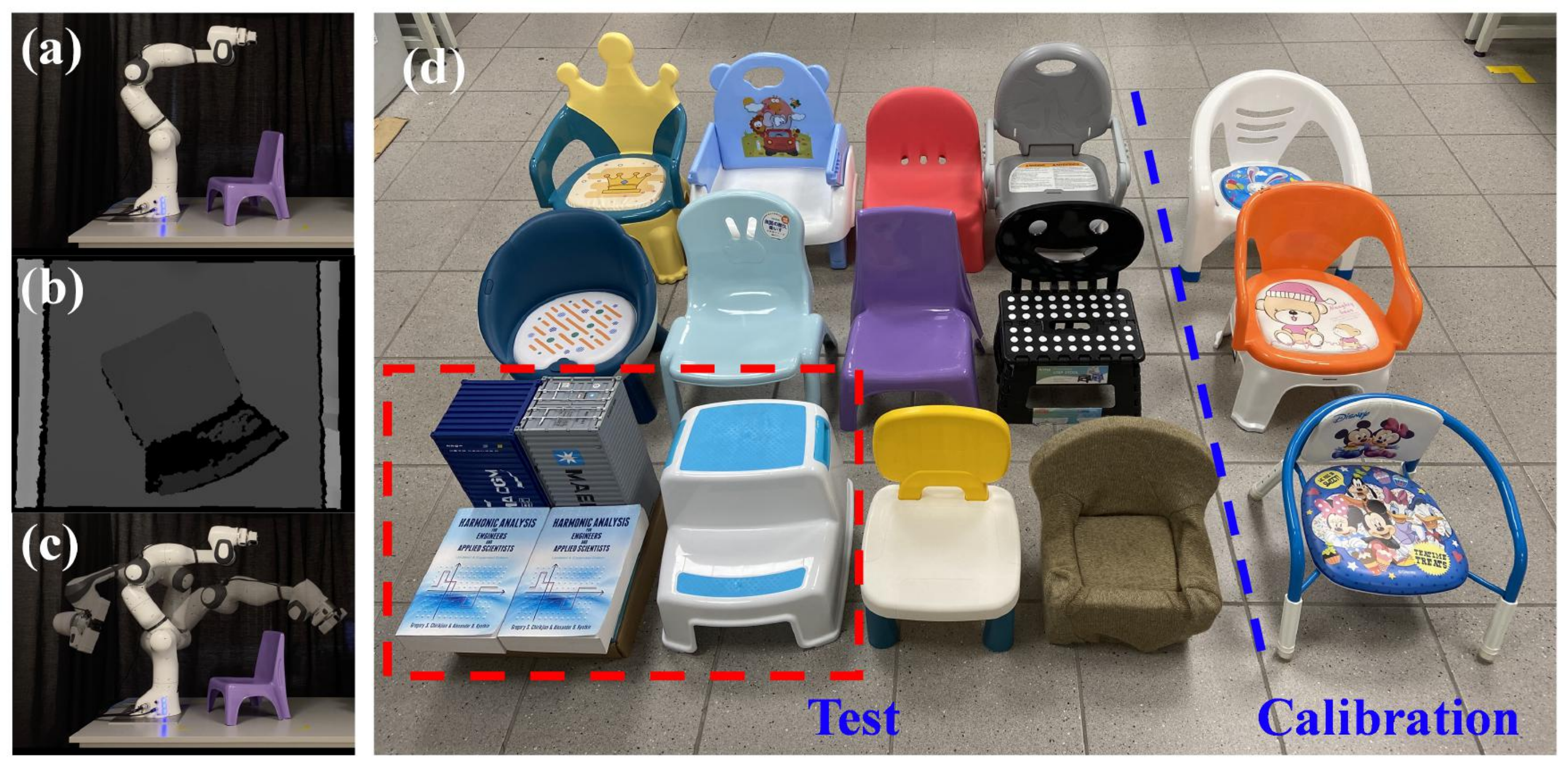}
    \caption{\textbf{Data.} 
    (b) shows a captured depth image when the robot is at the capturing pose shown in (a). 
    (c) shows three capturing poses of the 3D scanning. 
    (d) shows all the chairs in the dataset.
    Among the chairs in the test set, two of them, indicated in the red box, are not typical chairs but are able to afford the sitting affordance.
    One is a step stool; the other is improvised by assembling books and boxes.
    }
    \label{fig:4}
    \vspace{-0.6cm}
\end{figure}

\subsection{3D Scanning}
\label{sec:scanning}
In the experiment, the chair is placed randomly in a 50$\times$50cm$^{2}$ squared area in front of the robot arm in its upright pose.
The robot arm moves to 9 pre-defined collision-free configurations to capture depth images of the scene (Fig. \ref{fig:4}(a)(b)(c)).
The pose of the camera at each view is obtained from the forward kinematics of the robot arm.
This allows us to use TSDF Fusion \cite{curless1996volumetric} to densely reconstruct the scene and the point cloud.
The chair is segmented from the scene by plane segmentation in the PCL library \cite{rusu20113d}.

\subsection{Data}
Our dataset contains 15 chairs with diverse shapes and appearances (Fig. \ref{fig:4}(d)).
They are designed for children aged from 0 to 3.
We choose small chairs because the size of the chair is restricted by the workspace of the NAO and the robot arm.
If the chair is too tall, the NAO will be too short to put the bear on it; if the chair is too large, the robot arm will not be able to scan it.
We use 3 chairs (calibration set) to calibrate the simulation and the motion planning and control.
The rest 12 chairs (test set), which are unseen by the robot, are used for testing.

\subsection{Physical Simulation}
Pybullet \cite{coumanspybullet} is used as the physics engine for sitting imagination.
The chair and the virtual humanoid agent are imported with the URDF files which specify the mass, COM, inertia matrix, friction coefficient, and joint properties.
The collision in the simulation is modelled as inelastic.
The physics attributes of the chair is computed with Meshlab \cite{cignoni2008meshlab}.
As the chairs in the dataset are designed for children, we set the height and weight of the agent in the simulation accordingly \cite{onis2008child}.

\section{RESULTS \& DISCUSSIONS}
\label{sec:results}
\begin{figure*}[!htb]
    \centering
    \includegraphics[width=1.95\columnwidth]{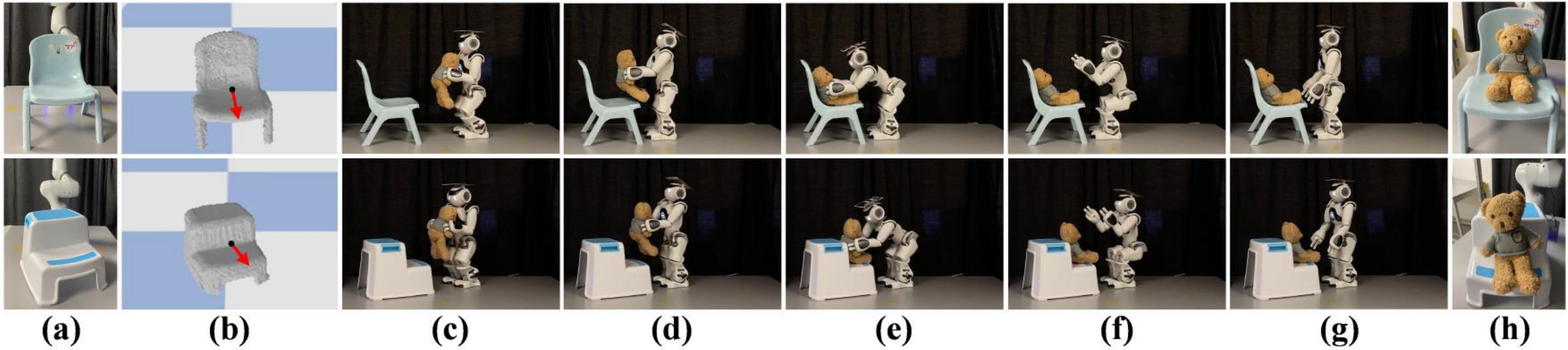}
    \caption{\textbf{Result: Accessible.}
    (a) Snapshot of the chair.
    (b) Imagined sitting pose.
    (c) The robot walks to the chair.
    (d) and (e) show the beginning and the end of the motion of seating the bear, respectively.
    (f) and (g) show the motion of retrieving hands and uprighting the body after seating the bear.
    (h) Results.}
    \label{fig:6}
    \vspace{-0.3cm}
\end{figure*}
\begin{figure*}[!h]
    \centering
    \includegraphics[width=1.95\columnwidth]{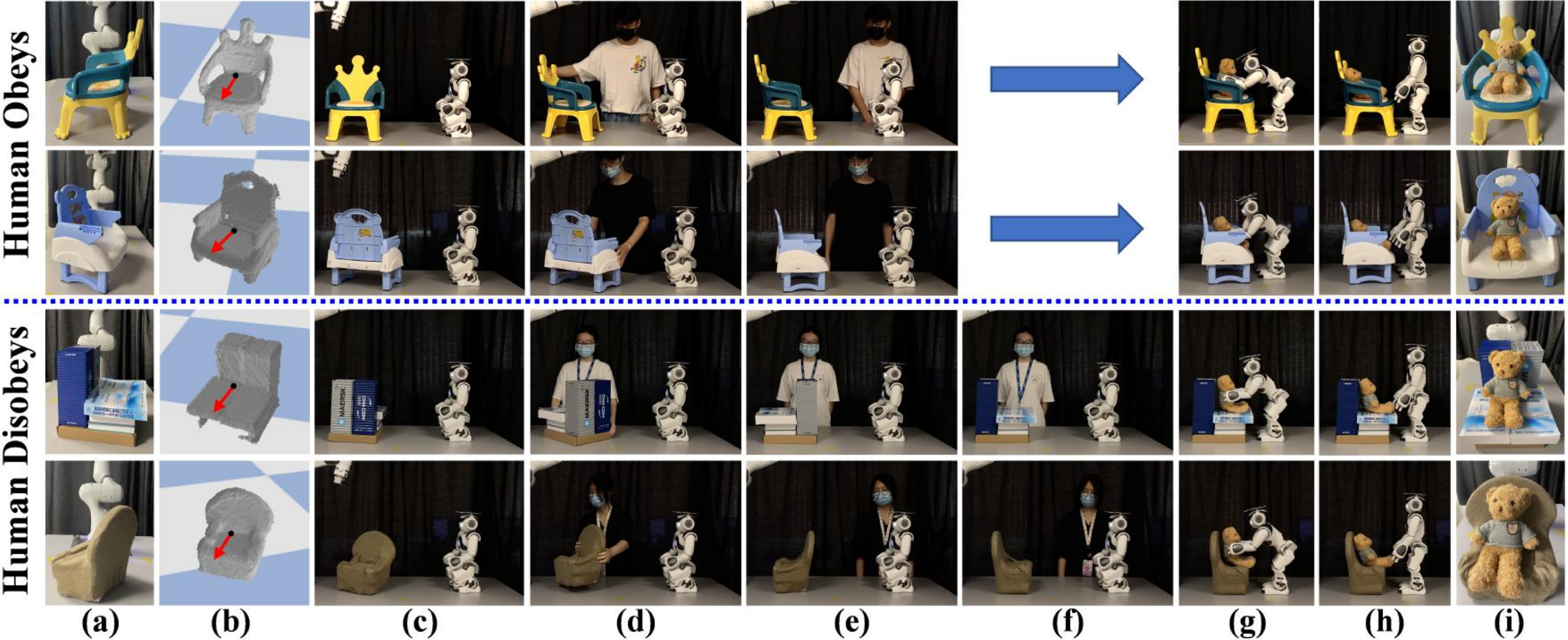}
    \caption{\textbf{Result: Inaccessible.}
    (a) Snapshot of the chair.
    (b) Imagined sitting pose.
    (c) Before human assistance.
    (d) The human obeys/disobeys the instruction given by the robot to rotate the chair.
    (e) After the first rotation. In the experiments of Human Obeys, the chair is already accessible after the first rotation.
    (f) After the last rotation.
    (g) and (h) show the beginning and the end of seating the bear on the chair, respectively.
    (i) Results.}
    \label{fig:7}
    \vspace{-0.5cm}
\end{figure*}

We implement our method with Robot Operating System (ROS) on a computer running Intel Core i9-10920X @ 3.5GHz CPU.
Our unoptimized implementation takes about 70s, 4s, and 29s for 3D scanning and reconstruction, sitting imagination, and motion planning, respectively.
The walking and seating motions take about 67s and 24s, respectively.

\textbf{Accessible.}
In the first set of experiments, the chair is placed such that it is accessible for seating the bear.
The sitting direction points towards the NAO with a deviation within a range of $\pm 30$ degrees.
No human assistance is needed in this case.
For each chair in the test set, we place it in 3 different poses for testing (36 trials in total) (Fig. \ref{fig:6}).

\textbf{Inaccessible + Human Obeys.}
In the second set of experiments, the chair is placed such that it is inaccessible.
The sitting direction of the chair either points towards 1) the robot arm or 2) to the edges of the planning arena (Fig. \ref{fig:7}(a)).
In both cases, no valid $SE(2)$ trajectories can be found in the initial configuration.
Human assistance is needed to rotate the chair and make it accessible.
We recruit 6 volunteers to participate in the experiments.
For each chair in the test set, we place it in 2 different inaccessible poses (24 trials in total).
We ask the human to obey the instruction given by the NAO throughout the whole trial (Fig. \ref{fig:7}).

\textbf{Inaccessible + Human Disobeys.}
There exist many uncertainties in human assistance (\textit{e.g.}, the human is distracted or misunderstands the instruction), resulting in inaccessibility even after a rotation.
In this set of experiments, we want to test the robustness of our method in addressing these uncertainties.
For each chair in the test set, we place it in an inaccessible pose as in the Inaccessible + Human Obeys setting (12 trials in total).
We ask the human to deliberately disobey the first instruction given by the NAO and obey the following instructions (Fig. \ref{fig:7}).

We recruit 15 annotators to annotate the experiment results.
Each trial is annotated by five different annotators.
For each trial, we show the experiment video and the images of the bear at the end of the trial.
The annotator is then asked 1) ``Do you think the robot has been successful in seating the bear on the chair?"
For the trial with human assistance, we also 2) ``Is the chair accessible after the human assistance?"
The results on the test set are shown in the Tab. \ref{tab:1}.

\begin{table}[]
\begin{center}
\begin{tabular}{p{3.6cm}p{0.8cm}<{\centering}p{0.65cm}<{\centering}p{0.65cm}<{\centering}p{0.65cm}<{\centering}}

\multicolumn{5}{l}{\textbf{(1) Seating Bear Success}}                                                                         \\
\toprule
\multicolumn{1}{c}{\multirow{2}{*}{Experiment}} & Trial                    & \multicolumn{3}{c}{Positive Annotation Num.} \\ \cline{3-5} 
\multicolumn{1}{c}{}                            & Num. & = 5 & $\geq$ 4 & $\geq$ 3\\ 
\midrule
Accessible                     & 36                   & 26    & 34   & 34   \\
Inaccessible + Human Obeys     & 24                   & 18    & 22   & 23   \\
Inaccessible + Human Disobeys  & 12                   & 10    & 10   & 11   \\
Total                          & 72                   & 54    & 66   & 68  \\
\bottomrule

\\
\multicolumn{5}{l}{\textbf{(2) Chair Accessible After Human Assistance}}                                                                 \\
\toprule
\multicolumn{1}{c}{\multirow{2}{*}{Experiment}} & Assist.                    & \multicolumn{3}{c}{Positive Annotation Num.} \\ \cline{3-5} 
\multicolumn{1}{c}{}                            & Num. & = 5 & $\geq$ 4 & $\geq$ 3\\
\midrule
Inaccessible + Human Obeys    & 23                   & 23      & 23      & 23     \\
Inaccessible + Human Disobeys & 11                   & 10      & 11      & 11     \\
Total                         & 34                   & 33      & 34      & 34     \\
\bottomrule
\end{tabular}
\caption{Experiment Results on the Test Set.}
\label{tab:1}
\end{center}
\vspace{-0.9cm}
\end{table}

In table (1) of Tab. \ref{tab:1}, we show the number of trials with at least 5, 4, and 3 positive annotations to the first question.
We count the trial as successful if the sitting imagination found the sitting pose \textit{and} more than half of all the 5 annotations (at least 3) are positive.
In general, we are able to achieve a very high success rate of $94.4\%$ on the 12 unseen chairs in the test set (68 out of 72).
The step stool accounts for all the 4 failure cases in the 72 trials.
The sitting imagination fails to find the sitting pose because the depth of the seat is too small.
A successful trial of the step stool is shown in Fig. \ref{fig:6}.
Remarkably, the success rate of all the 6 trials of the \textit{improvised} chair is $100\%$.
This opens up a promising potential of our imagination method to discover the affordance of an object which can afford the sitting functionality despite it not being a typical chair.

In table (2), we show the total number of trials with human assistance and the number of trials with at least 5, 4, and 3 positive annotations to the second question.
The goal of human assistance is to make the chair accessible for placing the bear.
We evaluate the accessibility of the chair at the end of human assistance.
We only count the trial in which human assistance is involved, \textit{i.e.}, we disregard the failure trials where the sitting imagination fails to find the sitting pose and no instructions will be given.
We consider the chair accessible if more than half of all the annotations are positive.
The results show that our human assistance module is very effective -- it achieves a $100\%$ success rate in making the chair accessible in all the trials with successful sitting imagination.

\section{CONCLUSIONS \& FUTURE WORK}
\label{sec:conclusions}
We propose a novel method for robots to imagine the sitting pose for a previously unseen chair.
We develop a robotic system which is able to seat a teddy bear on the chair autonomously via robot imagination.
Moreover, we introduce a module for robots to ask a human to assist with changing the accessibility of the chair when it is in an inaccessible pose.
Results show that our method enables the robot to seat the bear on 12 previously unseen chairs in 72 trials with a very high success rate.
The human assistance module is also shown to be very effective in making the chair accessible from inaccessible poses.
Future work can focus on exploring more versatile whole-body motion planning methods for seating the bear on chairs.





\section*{ACKNOWLEDGMENT}
The authors thank Yuanfeng Han for discussions, all the volunteers for the human assistance in the experiments, and all the annotators for the human annotations.

\bibliographystyle{IEEEtran}
\bibliography{references}

\end{document}